# Deriving the Scaled-Dot-Function via Maximum Likelihood Estimation and Maximum Entropy Approach

Jiyong Ma

Oracle Corporation

## Abstract

In this paper, we present a maximum likelihood estimation approach to determine the value vector in transformer models. We model the sequence of value vectors, key vectors, and the query vector as a sequence of Gaussian distributions. The variance in each Gaussian distribution depends on the time step, the corresponding key vector, and the query vector. The mean value in each Gaussian distribution depends on the time step, and the corresponding value vector. This analysis may offer a new explanation of the scaled-dot-product function or softmax function used in transformer architectures [1]. Another explanation, inspired by [4], is based on the maximum entropy approach in natural language processing [5]. In this approach, a query vector and key vectors are used to derive the feature functions for the maximum entropy model.

## Estimation of Softmax Function

In this discussion, we focus on the self-attention mechanism employed within the encoder or decoder components of transformer models [1].

We assume that $\boldsymbol{q}, \boldsymbol{v}, \boldsymbol{k}$ are d-dimensional random vectors respectively, where $\boldsymbol{q}$ is a query vector, $\boldsymbol{v}$ is a value vector, and $\boldsymbol{k}$ is a key vector. The vectors $\boldsymbol{v}$, $\boldsymbol{q}$, and $\boldsymbol{k}$ all reside in d-dimensional space $\boldsymbol{R}^d$ respectively.

The vectors $\boldsymbol{q}$ and $\boldsymbol{v}$ are projections of a high-dimensional input vector $\boldsymbol{x}$, which lies in m-dimensional space $\boldsymbol{R}^m$. The dimension $m$ is much larger than the dimension $d$. The space $\boldsymbol{R}^d$ is a d-dimensional subspace within the larger space $\boldsymbol{R}^m$.

For simplicity, we will drop the dependent variable $\boldsymbol{x}$ in vectors $\boldsymbol{v}$ and $\boldsymbol{q}$ in the following discussion.

Let $T$ denote the sequence length. We denote a sequence of query vectors as $\{\boldsymbol{q}_1, \boldsymbol{q}_2, \boldsymbol{q}_3, \ldots, \boldsymbol{q}_T\}$. Similarly, we denote a sequence of value vectors as $\{\boldsymbol{v}_1, \boldsymbol{v}_2, \boldsymbol{v}_3, \ldots, \boldsymbol{v}_T\}$.

The vectors $\boldsymbol{q}_i$ and $\boldsymbol{v}_i$ are two low-dimensional projections of an input vector $\boldsymbol{x}_i$. The vector $\boldsymbol{x}_i$ lives in m-dimensional space $\boldsymbol{R}^m$.

We also have a sequence of key vectors denoted as $\{k_1, k_2, k_3, \ldots, k_T\}$. Each key vector $k_i$ represents an address memory unit, which is also a low-dimensional projection of the input vector $x_i$. Each key vector $k_i$ can also be viewed as a hidden state associated a value vector $v_i$ which represents a content or feature memory unit.

A query vector $q$ will match each address memory unit $k_i$ to compute a similarity score. The similarity score can be represented as the cross-correlation value, which is the inner product of the two vectors $q$ and $k_i$ denoted as $q^t k_i$. For the definition of inner product of two vectors, please refer to Appendix A. Note that the vector $q$ is one of vectors in the sequence $\{q_1, q_2, q_3, \ldots, q_T\}$.

A schematic diagram depicting the self-attention layer in transformer models is shown in Fig. 1.

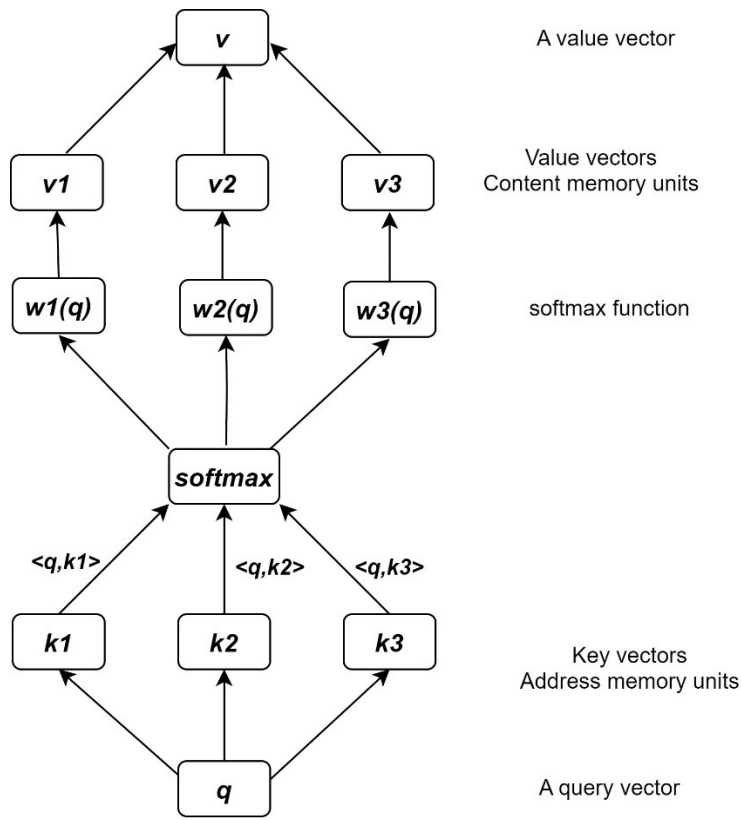

Figure 1. *The diagram depicts the self-attention layer in a transformer model. The input is a query vector **q**, represented as the node at the bottom of the diagram. The output is a value vector **v**, shown as the node at the top. In this example, the sequence length is 3. The key vectors are denoted as **k1, k2, k3**, and the corresponding value vectors are **v1, v2, v3**. The symbols <q, k1>, <q, k2>, and <q, k3> denote the inner products between the query vector **q** and vectors **k1, k2, k3** respectively. w1(q), w2(q), w3(q) are the outputs of the softmax function. The process starts from a query vector **q** being sent to the address memory units represented by the key vectors **k1, k2, k3**. These units compute the inner products between the query and key vectors. A inter product value represent similarity score between the query and the key. These similarity scores pass thought a softmax function, producing the weighting values w1(q), w2(q), w3(q).*

*Finally, the output value vector **v** is computed as the weighted summation of the value vectors **v1, v2, v3**, using softmax weights values w1(q), w2(q), w3(q) as coefficients.*

This self-attention mechanism allows the model to dynamically attend to different parts of the input sequence when computing the output value vector.

The definition of the softmax weighting function is described in [1]:

$$w_i(\boldsymbol{q}) = \frac{exp(\alpha \boldsymbol{q}^t \boldsymbol{k}_i)}{\sum_{j=1}^{T} exp(\alpha \boldsymbol{q}^t \boldsymbol{k}_j)} \tag{1}$$

Where:

- The notation exp(y) is used to represent the exponential function $e^y$.
- Where e is the base of the natural logarithm, approximately equal to 2.71828.
- $\alpha$ is a positive scaling factor.
- *t* is a transpose operator, which transforms a column vector to a row vector.
- $\boldsymbol{q}^t \boldsymbol{k}_i$ or the dot product denoted as <$\boldsymbol{q}, \boldsymbol{k}_i$> is the inner product between the query vector $\boldsymbol{q}$ and i-th key vector $\boldsymbol{k}_i$.
- The denominator $\sum_{j=1}^{T} exp(\alpha \boldsymbol{q}^t \boldsymbol{k}_j)$ is the sum of the exponentials of all the inner products between the query $\boldsymbol{q}$ and the set of key vectors $\{\boldsymbol{k}_1, \boldsymbol{k}_2, \boldsymbol{k}_3, \ldots, \boldsymbol{k}_T\}$.
- $w_i(\boldsymbol{q})$ is the softmax weight corresponding to the i-th key vector $\boldsymbol{k}_i$

The inclusion of this scaling factor $\alpha$ allows the model to control the sharpness of the softmax weights. A larger value of α will make the softmax weights more concentrated on the most similar key vectors, while a smaller value of α will result in a more uniform distribution of weights.

This softmax function takes the inner product similarities between the query $\boldsymbol{q}$ and each key vector $\boldsymbol{k}_i$, applies an exponential to them, and then normalizes them to produce a set of positive weights $w_i(\boldsymbol{q})$ that sum to 1.

The softmax function has the effect of amplifying the weights of key vectors that are most similar to the query vector, while suppressing the weights of less similar key vectors. This allows the self-attention mechanism to focus on the most relevant parts of the input sequence when computing the output value vector.

The main goal of the upcoming discussion is to derive the softmax function by starting from a maximum likelihood estimation of a joint probabilistic distribution. This is a very interesting approach, as it can provide a deeper probabilistic interpretation of the softmax function used in self-attention.

By deriving the softmax function from the principle of maximum likelihood estimation, we can potentially gain a better understanding of the underlying statistical assumptions and properties of the softmax mechanism. This could lead to further insights or improvements in how the attention weights are computed in transformer models.

## Maximum Likelihood Estimation

Inspired by probabilistic transformers [2], we propose another way to derive the softmax function in transformers [1]. We assume all components $v_{ij}$ in the $\boldsymbol{v_i} = (v_{i1}, v_{i2}, \cdots, v_{id})^t$ are independent random variables. For the simplicity and without loss of generality, we drop the coordinate index $j$ in the following discussion. The scale value $v_{ij}$ is denoted as $v_{i*}$.

The Gaussian distribution of one of coordinate $v_{i*}$ in the vector $\boldsymbol{v_i} = (v_{i1}, v_{i2}, \cdots, v_{id})^t$ at time $i$ is denoted as follows:

$$g_i(v_{i*}|v, \boldsymbol{q}) = \left(\frac{\theta(i, \boldsymbol{q})\beta}{2\pi}\right)^{1/2} exp(-\frac{\theta(i, \boldsymbol{q})\beta}{2}(v - v_{i*})^2) \tag{2}$$

$$\theta(i, \boldsymbol{q}) = exp(\alpha \boldsymbol{q}^t \boldsymbol{k_i}) \tag{3}$$

Where:

- $v$ is one coordinate of the vector $\boldsymbol{v}$, which is defined as $\boldsymbol{v} = (v_1, v_2, \cdots, v_d)^t$.
- $v$ is the mean value of the Gaussian distribution.
- $v_{i*}$ is one coordinate of the vector $\boldsymbol{v_i}$, which is defined as $\boldsymbol{v_i} = (v_{i1}, v_{i2}, \cdots, v_{id})^t$.
- $\alpha$ and $\beta$ are two positive constants.
- $\sigma^2(i, \boldsymbol{q}) = 1/(\theta(i, \boldsymbol{q})\beta) = exp(-\alpha \boldsymbol{q}^t \boldsymbol{k_i})/\beta$ represents of the variance of the Gaussian distribution.
- The variances of different coordinates are the same $\sigma^2(i, \boldsymbol{q})$, which doesn't depend on the index coordinate index $i$.
- The mean value of the Gaussian distribution is $v_{i*}$.

Note the notation difference between the vector $\boldsymbol{v}$ and its one coordinate scale value $v$: the vector is in bold font.

Based on the previous definition of the time-dependent multivariate Gaussian distribution over $v_{i*}$, the joint probability density of the sequence of key and value vectors from time 1 to $T$, along with the query vector, can be written as:

$$f(v_{i*}|v, \boldsymbol{q}) = \prod_{i=1}^{T} g_i(v_{i*}|v, \boldsymbol{q}) \tag{4}$$

The joint probability is simply the product of the individual Gaussian densities at each time step $i$, under the assumption of independence between the time steps. We didn't include all components in the vector $\boldsymbol{v}$, because the variances of all components are the same.

This joint probability expression is the starting point for the maximum likelihood estimation and derivation of the softmax function.

The next step is to describe how to estimate the value $v$ using this probabilistic framework.

Let's proceed with defining the log-likelihood function based on the joint probability expression in Eq. 4.

The log-likelihood function is defined as:

$$L = \log(f(v_{i*}|v, \boldsymbol{q}))$$

We can express the log-likelihood as the sum of the log-probabilities of the individual Gaussian distributions at each time step $i$. Substitute Eq.4 into the above equation, we have the following expanded equation:

$$L = \log(f(v_{i*}|v, \boldsymbol{q})) = \sum_{i=1}^{T} \log(g_i(v_{i*}|v, \boldsymbol{q}))$$

The goal is to maximize this log-likelihood function $L$ with respect to the unknown parameter $v$, in order to obtain the maximum likelihood estimate. This would involve taking the derivative of the log-likelihood with respect to $v$, and setting them equal to 0 to find the optimal value.

From the above equation, let's take the partial derivative of the log-likelihood function $L$ with respect to $v$:

$$\frac{\partial L}{\partial v} = \sum_{i=1}^{T} \frac{\partial \log(g_i(v_{i*}|v, \boldsymbol{q}))}{\partial v} \tag{5}$$

This partial derivative expression provides the gradient of the log-likelihood $L$ with respect to $v$.

Substitute Eq. 2 into the above equation, we have:

$$\frac{\partial \log(g_i(v_{i*}|v, \boldsymbol{q}))}{\partial v} = -\beta \theta(i, \boldsymbol{q})(v - v_{i*}) \tag{6}$$

The detailed description of the derivation of the right-hand side of the previous equation is shown in the Appendix A.

Substitute of the previous equation into Eq. 5, and let the partial derivatives equal to zero, i.e., $\frac{\partial L}{\partial v} = 0$, we have:

$$\frac{\partial L}{\partial v} = \sum_{i=1}^{T} \frac{\partial \log(g_i(v_{i*}|v, \boldsymbol{q}))}{\partial v} = \sum_{i=1}^{T} -\beta \theta(i, \boldsymbol{q})(v - v_{i*}) = 0 \tag{7}$$

From the above equation, we have:

$$\sum_{i=1}^{T} \theta(i,q)(v - v_{i*}) = v \sum_{i=1}^{T} \theta(i,q) - \sum_{i=1}^{T} \theta(i,q)v_{i*} = 0 \tag{8}$$

From the above equation, we have the following equation to estimate $v$:

$$v = \sum_{i=1}^{T} \theta(i,q)v_{i*} / \sum_{i=1}^{T} \theta(i,q) \tag{9}$$

From Eq. 9, we have the following weighting function denoted as $p_i(q)$:

$$p_i(q) = \theta(i,q) / \sum_{j=1}^{T} \theta(j,q) \tag{10}$$

Substitute Eq. 3 into the above equation, we can get Eq.1. Note that $p_i(q) = w_i(q)$, from Eq.9, we have the following equation:

$$v = \sum_{i=1}^{T} w_i(q) v_{i*} \tag{11}$$

Note that $exp(\alpha q^t k_i)$ represents the inverse of variance $\sigma^2(i,q)$ defined in the following equation:

$$\theta(i,q) = exp(\alpha q^t k_i) = \frac{1}{\sigma^2(i,q)\beta} \tag{12}$$

From the above equation, we know the softmax function in Eq. 1 is a weighting function of the inverse of variance $\sigma^2(i,q) = exp(-\alpha q^t k_i)/\beta$, which depends on time $i$, a query vector $q$ and a key vector $k_i$. Note the constant $\beta$ can be canceled in the nominator and the denominator.

The weighted value vector $v$ in Eq. 9 is also called as context vector in [3]. The weighting factors $w_i(q)$ of the softmax function in Eq.1 are used to select the relevant context/value vectors in $\{v_1, v_2, v_3, \ldots, v_T\}$.

Based on the definition of the weighting function $w_i(q)$, we have the following key points:

- The weighting factors $w_i(q)$ represent the contribution of each value vector $v_i$ to the final output, given the query vector $q$.
- These weighting factors may be sparse, meaning that only a small number of the value vectors are highly relevant to the query $q$.

This sparsity property is an important characteristic, as it allows the self-attention mechanism to focus on the most relevant parts of the input sequence when computing the output. By assigning higher weights to the most salient value vectors, the model can selectively attend to the most informative parts of the input.

The sparsity of the weighting factors $w_i(q)$ is likely achieved through the use of the Gaussian modeling and the inverse $\sigma^2(i, q) = exp(-\alpha q^t k_i)/\beta$ in the definition of $w_i(q)$. This can have the effect of suppressing the weights of key vectors that are less similar to the query $q$.

As to time complexity to compute the weighting factors, the key points are as follows:

- For each time step $i$ from $1$ to $T$, each key vector $k_i$ in the sequence $\{k_1, k_2, k_3, \ldots, k_T\}$ has an associated value vector $v_i$.
- The weighting factor $w_i(q)$ represents the contribution of the i-th value vector $v_i$ to the final output, given the query vector $q$.
- The time complexity for computing the weighting factors $w_i(q)$ for i = 1 to $T$ is O($T$).
- The time complexity for computing all weighting factors $w_i(q_j)$ for i = 1 to $T$ and for j = 1 to T is O($T*T$), resulting in a quadratic dependence on the sequence length $T$.

## Maximum Entropy Approach

Another way to derive the softmax weighting function defined in Eq. 1 is based on the maximum entropy approach in natural language processing [5].

The feature functions are defined as $q^t k_i$ or the dot product denoted as $<q, k_i>$, which is the inner product between the query vector $q$ and i-th key vector $k_i$.

A sequence of key vectors is denoted as $\{k_1, k_2, k_3, \ldots, k_T\}$. Each key vector $k_i$ represents an address memory unit.

A query vector $q$ matches each address memory unit $k_i$ to compute the feature function score.

The variable $y$ of the feature function $f_i(x, y)$ defined in the Eq. 10 in [5] is the key index, with values ranging from 1 to $T$. while $x$ is the query vector $q$. The conditional probability is as follows:

$$p(y|x) = \frac{1}{Z_\lambda(x)} exp(\sum_i \lambda_i f_i(x, y)) \tag{13}$$

$$Z_\lambda(x) = \sum_{y=1}^{T} exp(\sum_i (\lambda_i f_i(x, y))) \tag{14}$$

$$f_i(x, y) = \begin{cases} q^t k_i, & if \quad y = i \\ 0, & if \quad y \neq i \end{cases} \tag{15}$$

Substitute Eq. 15 into Eq. 13, and Eq. 14 we have,

$$p(i|q) = \frac{exp(\lambda_i q^t k_i)}{\sum_{j=1}^{T} exp(\lambda_j q^t k_j))} \tag{16}$$

When $\lambda_i$ equals to $\alpha$, the positive scaling factor, Eq. 16 is the same as Eq. 1

## Conclusion

In this paper, we presented a maximum likelihood estimation approach to the derive scaled-dot-product function used in transformer models. The main steps were:

1. We modeled the concatenated vector of the query vector $q$ and the value vector $v$ as a time-dependent multivariate Gaussian distribution.
2. We then derived the partial derivative of the log-likelihood function with respect to the value vector $v$, which led to a weighting function $w_i(q)$.
3. These weighting factors $w_i(q)$ capture the relevance of each key vector to the given query vector $q$.
4. The computation of these weighting factors has a quadratic time complexity $O(T*T)$, which can be computationally expensive for long input sequences.

The maximum likelihood derivation provides a principled probabilistic interpretation of the softmax function used in transformer self-attention mechanisms. This analysis may lead to a better understanding of the underlying statistical assumptions and properties of the attention weights computed by transformer models.

We also presented a maximum entropy approach to derive the scaled-dot-product function. In this approach, a query vector and key vectors are used to derive the feature functions used in maximum entropy approach.

## Appendix A

We use the following notations: A column vector $x = (x_1, x_2, \cdots, x_m)^t$, where $t$ is the transpose operator. A column vector $y = (y_1, y_2, \cdots, y_m)^t$.

The multiplication of real numbers $a_1, a_2, \cdots, a_N$ is denoted as:

$$\prod_{i=1}^{N} a_i = a_1 a_2 \cdots a_N$$

The summation of real numbers $a_1, a_2, \cdots, a_N$ is denoted as:

$$\sum_{i=1}^{N} a_i = a_1 + a_2 + \cdots + a_N$$

The inner product of two vector $x$ and $y$ is defined as the following equation:

$$< x, y > = x^t y = y^t x = \sum_{i=1}^{m} x_i y_i = x_1 y_1 + x_2 y_2 + \cdots + x_m y_m \quad (a1)$$

When *x=y,* from the above equation, we have:

$$\|x\|^2 = x^t x = \sum_{i=1}^{m} x_i^2 \quad (a2)$$

From the above equation, we have the length of the vector *x* as follows:

$$\|x\| = \sqrt{x^t x}$$

The cosine value of the angle $\vartheta$ between the vector *x* and *y* is as follows:

$$cos(\vartheta) = \frac{x^t y}{\|x\|\|y\|}$$

A softmax function is defined as:

$$softmax(x) = \frac{1}{\sum_{j=1}^{m} exp(x_j)} (exp(x_1), exp(x_2), exp(x_3), \cdots exp(x_m))$$

Note that Eq. 2 is defined as:

$$g_i(v_{i*}|v, q) = \left(\frac{\theta(i, q)\beta}{2\pi}\right)^{1/2} exp\left(-\frac{\theta(i, q)\beta}{2}(v - v_{i*})^2\right)$$

From the above equation we have:

$$log(v_{i*}|g_i(v, q)) = log\left(\left(\frac{\theta(i, q)\beta}{2\pi}\right)^{1/2} exp\left(-\frac{\theta(i, q)\beta}{2}(v - v_{i*})^2\right)\right)$$

$$log(v_{i*}|g_i(v, q)) = 1/2 log\left(\frac{\theta(i, q)\beta}{2\pi}\right) - \frac{\theta(i, q)\beta}{2}(v - v_{i*})^2$$

The first term in the right-hand side of the above equation doesn't depend on $v$, the partial derivative of the above equation is:

$$\frac{\partial log(v_{i*}|g_i(v, q))}{\partial v} = -\frac{\theta(i, q)\beta}{2} \frac{\partial (v - v_{i*})^2}{\partial v} = -\beta \theta(i, q)(v - v_{i*})$$

The above equation is the same as Eq. 6.